\documentclass[11pt,a4paper]{article}
\usepackage[]{emnlp2021}
\usepackage{times}
\usepackage{latexsym}

\usepackage{microtype}

\usepackage{booktabs} 

\usepackage{textcomp}
\usepackage{latexsym}
\usepackage{graphicx} 
\usepackage{microtype}
\usepackage{pifont}
\usepackage[font=small]{caption}
\usepackage[]{subcaption}
\usepackage{verbatim}
\usepackage{framed}
\usepackage{color, colortbl}
\usepackage[colorinlistoftodos]{todonotes}
\usepackage{amssymb}
\usepackage{amsmath}
\usepackage{breqn}
\usepackage{capt-of}
\usepackage{multirow}
\usepackage{algorithmicx}
\usepackage{algorithm}
\usepackage{algcompatible}
\usepackage[normalem]{ulem}
\usepackage{paralist}
\usepackage{enumitem}
\usepackage{balance}
\usepackage{tabularx}
\usepackage[all]{nowidow}

\definecolor{high}{HTML}{FFCE8E}

\def\aclpaperid{2716} 

\title{LMSOC: An Approach for Socially Sensitive Pretraining}

\author{Vivek Kulkarni \\
    Twitter Cortex\\ 
  {\tt vkulkarni@twitter.com} \\ \And
  Shubhanshu Mishra \\
  Twitter Cortex \\ 
  {\tt smishra@twitter.com} \\ \AND
  Aria Haghighi\\
  Twitter Cortex \\ 
  {\tt ahaghighi@twitter.com} \\
}

\newif\ifreviewer

\reviewerfalse

\ifreviewer 
\newcommand{\meta}[1]{\textcolor{brown}{$_{meta}$[#1]}}
\newcommand{\vivek}[1]{\textcolor{cyan}{$_{vivek}$[#1]}}
\newcommand{\aria}[1]{\textcolor{red}{$_{aria}$[#1]}}
\newcommand{\shubh}[1]{\textcolor{magenta}{$_{shubh}$[#1]}}

\else
\newcommand{\meta}[1]{\textcolor{brown}{}}
\newcommand{\vivek}[1]{\textcolor{cyan}{}}
\newcommand{\aria}[1]{\textcolor{red}{}}
\newcommand{\shubh}[1]{\textcolor{magenta}{}}

\fi

\newcommand{\ourmethod}{\textsc{Lmsoc}}
\newcommand{\ctrlmethod}{\textsc{Lmctrl}}
\newcommand{\bertmethod}{\textsc{Bert}}

\allowdisplaybreaks
\DeclareCaptionFont{10pt}{\fontsize{10pt}{12pt}\selectfont}
\date{}

\begin{document}
\maketitle
\begin{abstract}
While large-scale pretrained language models have been shown to learn effective linguistic representations for many NLP tasks, there remain many real-world contextual aspects of language that current approaches do not capture. For instance, consider a  cloze-test ``I enjoyed the \rule{0.75cm}{0.15mm} game this weekend'': the correct answer depends heavily on where the speaker is from, when the utterance occurred, and the speaker's broader social milieu and preferences.
Although language depends heavily on the geographical, temporal, and other social contexts of the speaker, these elements have not been incorporated into modern transformer-based language models. We propose a simple but effective approach to incorporate speaker social context into the learned representations of large-scale language models. Our method first learns dense representations of social contexts using graph representation learning algorithms and then primes language model pretraining with these social context representations. We evaluate our approach on geographically-sensitive language-modeling tasks and show a substantial improvement (more than 100\% relative lift on MRR) compared to baselines\footnote{Code is available at \url{https://github.com/twitter-research/lmsoc}.}.
\end{abstract}

\section{Introduction}

Language models are at the very heart of many modern NLP systems and applications \cite{young2018recent}. Representations derived from large-scale language models are used widely in many downstream NLP models \cite{peters2018deep, devlin-etal-2019-bert}. However, an implicit assumption made in most modern NLP systems (including language models) is that language is independent of extra-linguistic context such as speaker/author identity and their social setting.  While this simplifying assumption has undoubtedly encouraged remarkable progress in modeling language, there is overwhelming evidence in socio-linguistics that language understanding is influenced by the social context in which language is \emph{grounded} \cite{nguyen2016computational,hovy2018social, mishra2018detecting, garten2019incorporating, flek-2020-returning, bender-koller-2020-climbing}. In fact, language use on social media where every utterance is grounded in a specific social context (like time, geography, social groups, communities) reinforces this often ignored aspect of language. When NLP applications ignore this social context, they may perform sub-optimally underscoring the need for a richer integration of social contexts into NLP models \cite{pavalanathan2015confounds,lynn2017human, zamani-etal-2018-residualized, lynn2019tweet, may-etal-2019-measuring,kurita-etal-2019-measuring, welch-etal-2020-compositional, hovy2021importance}.

Prior attempts to better leverage the social context surrounding language while learning language representations have mostly focused on learning social context dependent word embeddings and have been primarily used to characterize language variation across many dimensions (time, geography, and demographics). These methods learn word embeddings for each specific social context and can capture how word meanings vary across these dimensions  \cite{bamman2014distributed,kulkarni2015statistically, hamilton2016diachronic,welch-etal-2020-compositional,welch-etal-2020-exploring}. However, word embedding based approaches in general suffer from two fundamental limitations: (a) word embeddings are not linguistically contextualized as noted by \citet{peters2018deep} (b) word embedding learning is transductive -- they can only generate embeddings for words observed during training and usually assume a finite word vocabulary and a set of social contexts all of which need to be seen during training. Recent approaches have addressed the first limitation by learning word representations that are contextualized by their  token-specific usage context \cite{peters2018deep,devlin-etal-2019-bert,liu2019roberta,yang2019xlnet,yang2019context}. The second limitation has been addressed by WordPiece tokenization  methods \cite{6289079, devlin-etal-2019-bert, liu2019roberta}. While these approaches have successfully captured linguistic context, they still do not capture social context in language representations.\footnote{Upon acceptance of this publication, we became aware of independent parallel work \citet{hofmann-etal-2021-dynamic} which attempts to learn word embeddings that are dynamic (depends on time etc.) and contextualized. In particular, \citet{hofmann-etal-2021-dynamic} change the architecture of BERT to replace the type-based word embedding lookup layer with an additive word embedding layer that adds temporal context dependent offset embeddings (that are learn-able) to the type-based embeddings. The full model is then trained with task-specific loss functions. In contrast, we introduce no new trainable parameters in our language model component, do not focus on the word embeddings themselves but on primarily enabling large scale language models to leverage social contexts of grounded language.}
``How can we learn linguistically contextualized and socially contextualized language representations?'' is the question we seek to answer in this paper.

 We propose \ourmethod\ to (a) learn representations of tokens that are both linguistically contextualized and socially sensitive and (b) enable the language model to inductively generate  representations for language grounded in social contexts it has never observed during the language model pre-training process. As an example, our model can enable  NLP systems to associate the right entity being referred to based on the broader user/social context in which an utterance like ``Our Prime Minister visited the UK last week.'' is grounded. 
 
 \begin{figure}[t!]
\centering
    \includegraphics[width=0.49\textwidth]{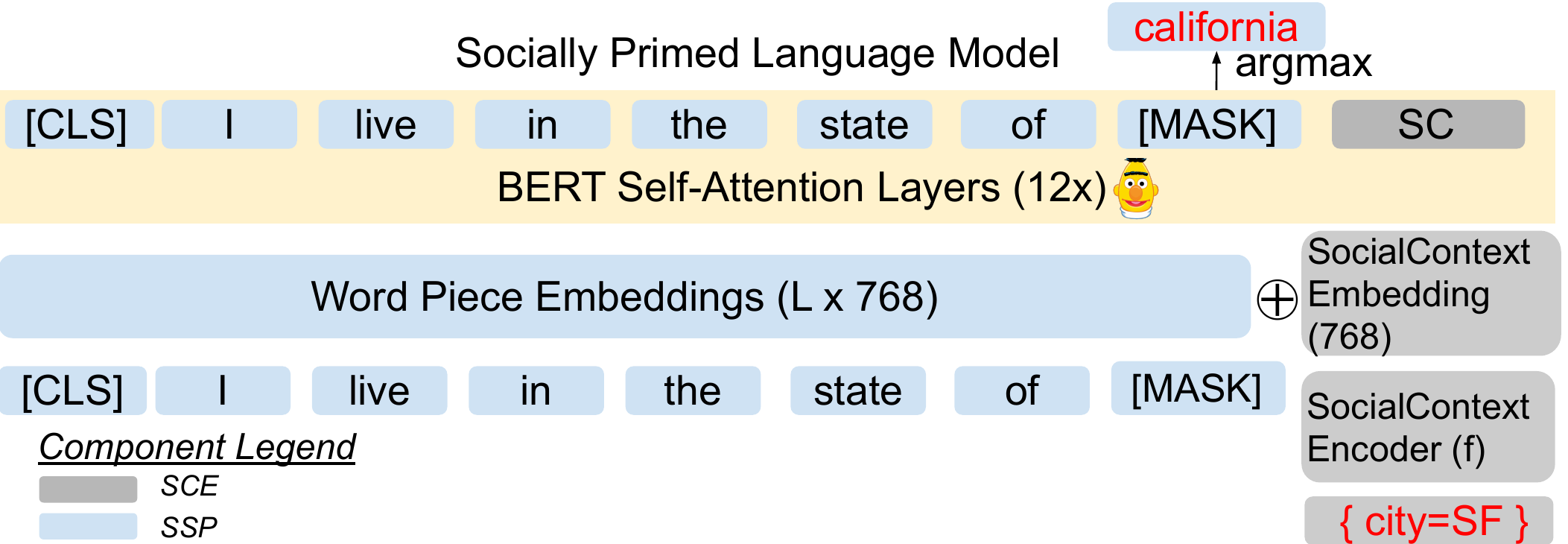}
	\caption{Overview of \ourmethod\ which has two components: a social context encoder (SCE) and a \textsc{Bert} based encoder for socially sensitive pre-training (SSP).}
	\label{fig:crown_jewel}
\end{figure}

\section{Model}
\label{sec:model}
\ourmethod\ has two components  (a) \textbf{SCE} -- a social context encoder and (b) \textbf{SSP} -- a standard \textsc{Bert} encoder altered to condition on the output of (a) (see Figure \ref{fig:crown_jewel}).

\paragraph{Social Context Encoder (SCE)} This component implements a function $f$ that maps a social context (like year, or location) to a $d$-dimensional embedding where similar social contexts are closer in this vector space than less similar ones. The specific method used to implement $f$ depends on the social context being modeled. Domain experts can choose to implement $f$ based on their expertise because the pre-trainer component is agnostic to how $f$ is implemented. One way of implementing $f$ is to encode the social contexts as a similarity network and use any graph representation learning algorithm to embed the nodes of this network in $\mathbb{R}^d$. Here, we use \textsc{Node2Vec} \cite{grover2016node2vec} as an expedient choice due to its simplicity and ease of training. Using this approach we show how to model commonly used social contexts like time and geographic location which we note fall under the \textsc{Context} category of the taxonomy of social factors outlined in \cite{hovy2021importance} -- a category that they observe can be quite challenging for NLP models to incorporate because of their overwhelmingly extra-linguistic nature. While in this work, we focus on just time and location, our method can also generalize to other social contexts (see Appendix \ref{sec:graph-rep}).

\paragraph{Socially Sensitive  Pretraining (SSP)} The second component is identical to a \textsc{Bert} encoder \cite{devlin-etal-2019-bert} with  a few modifications. First,  the social context representation obtained from the social context encoder is also incorporated to influence the representations of language learned when pre-training on the standard masked language modeling task. Specifically, let the sequence of input text tokens be $T = \langle w_1, w_2, w_3,\cdots w_n\rangle$ and the associated social context be $\texttt{SC} \in \mathbb{R}^d$. Note that standard \textsc{Bert} in its initial layers maps $T$ to a sequence of word piece embeddings denoted by $Q=\langle \Phi(q_1), \cdots \Phi(q_n)\rangle, \Phi(q_i)\in \mathbb{R}^d$ which are then transformed by higher layers. To incorporate the associated social context, we simply append $\texttt{SC}$ to $Q$ to yield $Q_{soc}=\langle \Phi(q_1), \cdots \Phi(q_n), \texttt{SC} \rangle$ which is then input to higher layers of \textsc{Bert}\footnote{We assume that the total length (including social context embedding) does not exceed the maximum length BERT’s architecture can handle.}. Second, we freeze $\texttt{SC}$ during training. These modifications enable further layers to attend to the social context and thus condition token representations on the social context in addition to the linguistic context. It is important to note the following: (a) Because the language model learns from a social context embedding, the language model can inductively yield representations of language grounded in social contexts that it has never observed in training. (b) No new trainable parameters are introduced in the language model component. This simple pre-training method thus learns representations of language that are contextualized both linguistically and socially.

\section{Evaluation}

\paragraph{Baseline Methods.} We evaluate the performance of \ourmethod\ against two baseline methods: (a) \bertmethod\  \cite{devlin-etal-2019-bert} which does not explicitly incorporate social context and (b) \ctrlmethod\ \cite{keskar2019ctrl} --  a very simple approach to incorporate social context into language models without altering the architecture of the language model itself. The key idea is to assign each social context a fixed code (a control code) \footnote{A control code is a distinctive name or number sequence.} which is appended to the input text. This approach has been shown to be useful for generating text conditioned on genre/domains \cite{keskar2019ctrl}. We adapt their approach but use \textsc{Bert} instead. While \ctrlmethod\  requires no change to the model architecture and conditions on the social context, this method cannot generalize to social contexts not seen during training (which we demonstrate empirically as well). Supporting new social contexts requires  the model to be retrained. 

\subsection{Evaluation on Synthetic Data}
\label{subsec:eval_synthetic}

We demonstrate the efficacy of \ourmethod\ on a cloze-test language modeling task using a synthetic corpus. This approach enables us to evaluate models in a very controlled setting, characterize their behavior, and demonstrate our method's face validity.  
 
\paragraph{Setup.} We consider a cloze-test language modeling task where the correct answer depends on the time (year) in which the sentence is grounded.  Noting that references to political positions in an utterance depend on the time period in which the utterance is grounded, we construct a synthetic corpus from two template sentences - (a) The president is [Name of President] and (b) The minister is [Name of minister] where each sentence is grounded in time.  Sentences grounded in year $t$ have the corresponding entity placeholder replaced with the name of the president (or minister) active in that specific year with active presidents/ministers changing every $5$ years. Our training data consists of $1000$ instances of each template sentence for each time point between the years $1900$ and $2000$ in steps of $5$ years. 

We evaluate all models on their ability to predict the correct token  replacing the masked token on test inputs of the template (``The [president/minister] of our country is [\textsc{Mask}]'', year), where we vary the year in which the sentence is grounded from $1900$ to $2000$. In particular, we  report the mean reciprocal rank (MRR) of the correct token over the test set. Note that this evaluation setting enables us to evaluate the performance of our model on social contexts not seen in training since the set of social contexts in evaluation is a super-set of those seen in training. To do well on this task, models need to leverage both the linguistic and the social context. Only using one or the other will result in sub-optimal performance\footnote{Notice that we also control for length of training sentences across social contexts in our controlled experiment since length could be a potential confounder.}.

To embed years, we use \textsc{node2vec} \cite{grover2016node2vec} on a simple linear chain graph where year $y$ is connected to $y-1$ and $y+1$. 

\begin{figure}[t!]
\centering
	\includegraphics[width=0.45\textwidth]{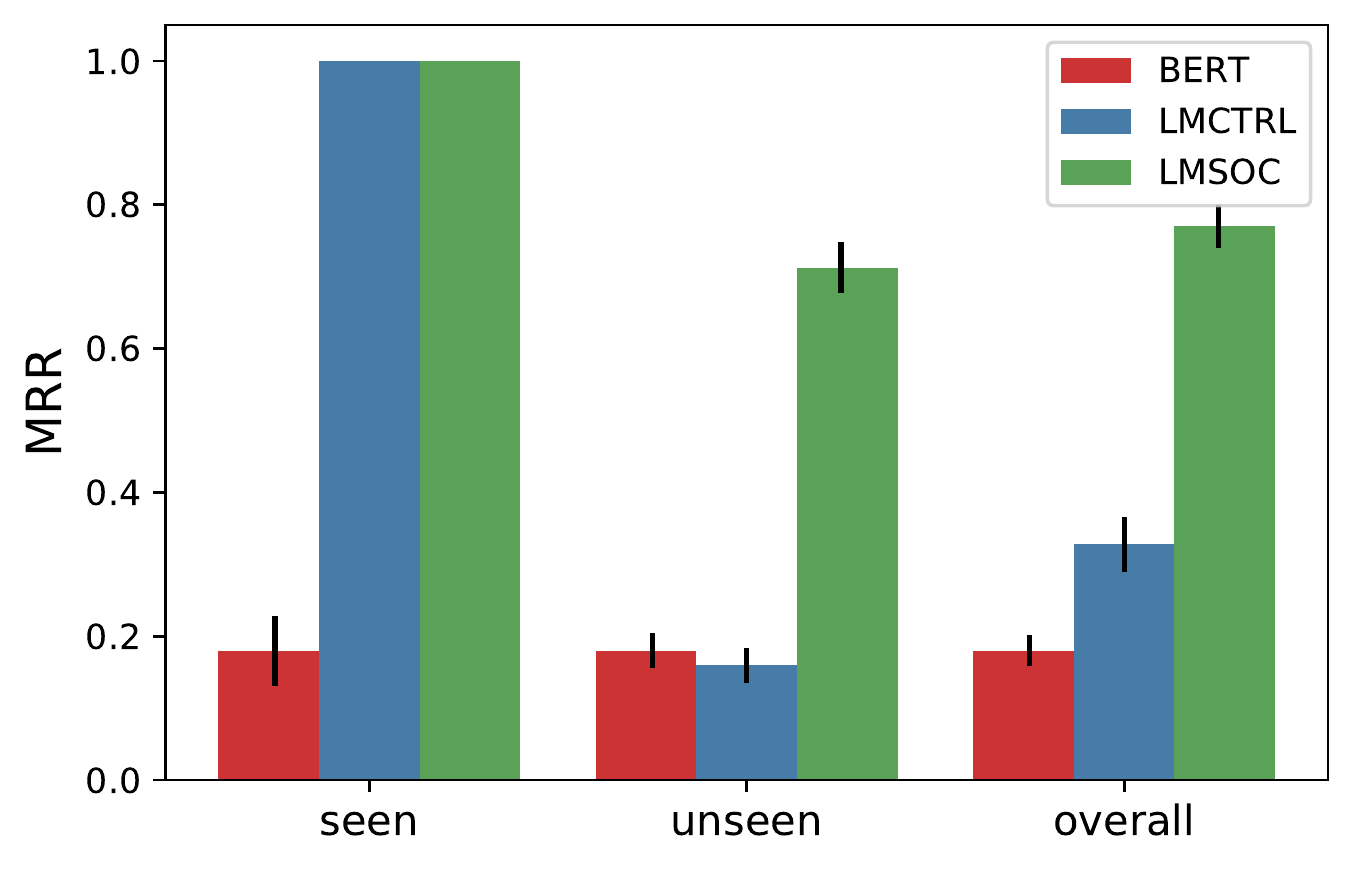}
	\caption{Performance of models on the synthetic data set as measured in terms of mean reciprocal rank (MRR, higher is better).  See Section \ref{subsec:eval_synthetic} for details.}
\label{fig:task1}
\end{figure}

\begin{table*}[htb!]
\small
\centering
\begin{tabular}{|l|cl|cl|}
\textbf{Model}  & \multicolumn{4}{c|}{\textbf{Task}}                                                         \\ \hline
        & \multicolumn{2}{c|}{\textbf{STATES}}              & \multicolumn{2}{c|}{\textbf{NFL}}                \\ 
        & \textbf{MRR $\uparrow$ ($95$\% CI)}     & \textbf{Mean Rank $\downarrow$ ($95$\% CI)} & \textbf{MRR $\uparrow$ ($95$ \% CI)}     & \textbf{Mean Rank $\downarrow$ ($95$\% CI)} \\ \hline
\bertmethod\   & $0.28 \ (0.20, 0.36)$ & $5.6 \ (4.17, 7.02)$    & $0.03 \ (0.02, 0.04)$ & $59.8 \ (47.1, 72.6)$   \\ \hline
\ctrlmethod\ & $0.41 \ (0.30, 0.51)$ & $9.8 \ (4.34, 15.29)$   & $0.03 \ (0.02, 0.04)$ & $86.8 \ (61.38, 112.2)$ \\ \hline
\ourmethod\  & $\mathbf{0.78\ (0.68, 0.89)}$ & $\mathbf{2.3 \ (0.72, 3.89)}$    & $\mathbf{0.15 \ (0.12, 0.19)}$ & $\mathbf{10.64 \ (6.66, 14.62)}$   \\ 
\end{tabular}
\caption{Overall performance of models on the \textbf{STATES} and \textbf{NFL} tasks using real world language data (including both seen and held-out social contexts) in terms of mean reciprocal rank (MRR, higher is better) and mean rank (lower is better). Our model \ourmethod\ outperforms all baselines significantly. See Section~\ref{subsubsec:real-eval-state} for more evaluation details.}
\label{tab:task2}
\end{table*}

\begin{table*}[htb!]
\small
\centering
\begin{tabular}{p{7cm}|l|p{6cm}}
\textbf{Input Sentence} & \textbf{Social Context} & \textbf{Top 10 predicted tokens} \\ \midrule \emph{I reside in the state of [MASK]} & \emph{San Diego} & \emph{\underline{california}, ca, texas, mexico} \\
\emph{I reside in the state of [MASK]} & \emph{Dallas} & \emph{\underline{texas}, houston, mexico, california, tx} \\
\emph{I reside in the state of [MASK]} & \emph{Tampa} & \emph{\underline{florida}, georgia, fl, texas, jacksonville} \\
\emph{The most popular nfl team in our state is [MASK]} &\emph{San Diego} & \emph{. the \underline{49ers} seattle patriots} \\ 
\emph{The most popular nfl team in our state is [MASK]} &\emph{Austin} & \emph{. alabama the … michigan florida atlanta \underline{texans} houston} \\ 
\end{tabular}
\caption{Top predictions of \ourmethod\ on sample instances grounded in unseen social contexts (expected tokens are underlined).}
\label{tab:task2_qualitative}
\end{table*}

\paragraph{Results.}
We present results for three settings in Figure \ref{fig:task1}: (a) Seen -- evaluation on held out test sentences but grounded in social contexts seen during training (b) Unseen -- evaluation on held out test sentences but grounded in social contexts unseen during training  (c) Overall -- combining both (a) and (b). First, note that \textsc{Bert} performs poorly in all settings as expected since it does not leverage the social context grounding the sentence. Next, observe that \textsc{Lmctrl} obtains perfect scores on the seen setting and significantly improves over the baseline overall. This is  because \textsc{Lmctrl} is able to condition on the social context. However it performs poorly when encountering unseen social contexts. This observation confirms that \textsc{Lmctrl} is able to learn representations that are dependent on social context, but requires all social contexts to be observed in training. Finally, our method \ourmethod\ significantly outperforms these baseline models in all settings, especially when evaluated on social contexts that are held out confirming the face validity of our model and suggests that our approach is effective at yielding representations that are both linguistically and socially contextualized.

\subsection{Evaluation on Real World Data}
\label{subsec:real-eval}

Here, we consider evaluating our model on real world language data. In the absence of standard benchmarks where predictions need to be conditioned on the broader social context, we consider the proxy task of geographically informed language modeling. Noting that correct answers to ``My hometown is [\textsc{Mask}]'' or ``We live in the state of [\textsc{Mask}]'' all depend on the geographical context that the utterance is grounded in, we consider a cloze language modeling evaluation comprising of three tasks (a) \textbf{STATES}: Recovering the geographical state that the author is likely referring to in an autobiographical sentence (b) \textbf{NFL}: Recovering the popular NFL (National Football League) teams that the author is most likely referring to in an utterance and (c) \textbf{CLOSECITY}: We evaluate the model's ability to align its predictions with geographical proximity between places. Note that the model has not been explicitly trained on these tasks. 

\paragraph{Data and Setup.} To construct our training data, we  obtain a random sample of $10$ million English tweets grounded in $10$ major US cities (each from a different state) as determined by the users' current location\footnote{The list of cities is available in the appendix.}. The social context associated with each tweet is this location.  

\subsubsection{STATES and NFL Tasks}
\label{subsubsec:real-eval-state} 
We evaluate our models on their performance at retrieving the correct entity for the two tasks using MRR of the expected answer in the model predictions. In both tasks, the test utterance may be grounded on a held out set of cities. For example, if the model was trained on tweets from Buffalo and San Francisco, then we may evaluate the model on its ability to predict the state being most likely referred to in the test sentence ``I reside in the state of [\textsc{Mask}]''. The correct answer is ``New York'' if the input is  grounded in Rochester and ``California'' if grounded in San Jose. In particular, we ground the input test sentence to one of the top 50 cities in the US by population. On the \textbf{STATES} task we use the test sentence ``We/I reside in the state of [\textsc{Mask}]'' whereas for the \textbf{NFL} task  we use ``The most popular NFL team in my state is [\textsc{Mask}].''\footnote{We obtained similar results for paraphrasings of these sentences.}

Finally, to embed cities we first construct a nearest neighbor graph ($k=5$) of cities based on pairwise geodesic distance computed using their geodesic co-ordinates and then embed the cities using \textsc{Node2Vec} on the constructed graph (see Appendix \ref{sec:graph-rep} for more details).

\subsubsection{CLOSECITY task}
\label{subsubsec:real-eval-closecity} 
To further evaluate the ability of the model to encode and leverage geographical proximity between places, we consider a task where we ask the model to predict plausible cities for the masked token in the following prompt: ``I drive to the city of [MASK] for work.'', where the utterance is grounded in a particular location akin to the \textbf{STATES} and \textbf{NFL} tasks. However, since there is no established ground truth for this task, we measure the geographical distance between the top predicted city/town of the model, and the input city (social context). Models that predict near-by cities or towns are better than models that predict far-away cities since one is more likely to drive to near-by cities for work than very far-away ones\footnote{If a surface form may link to multiple real-world locations, we give the models the benefit of doubt and assume they meant the closer location.}. Note that to ensure non-triviality, we exclude the input city as a valid candidate (or answer). Also, it is important to note that (a) the model is free to predict any city/town and (b) highly scoring answers do not necessarily correspond to largest cities in the input location's state or even cities in the same state. For example, if the input social-context is ``Buffalo, NY'', a model that predicts ``Toronto, Canada'' ($100$ km apart) is better than one that predicts ``New York City, NY'' ($470$ km apart). Aside from these differences, the rest of the setup is similar to the \textbf{STATES} and \textbf{NFL} tasks.

\begin{figure}[htb!]
\centering
	\includegraphics[width=0.45\textwidth]{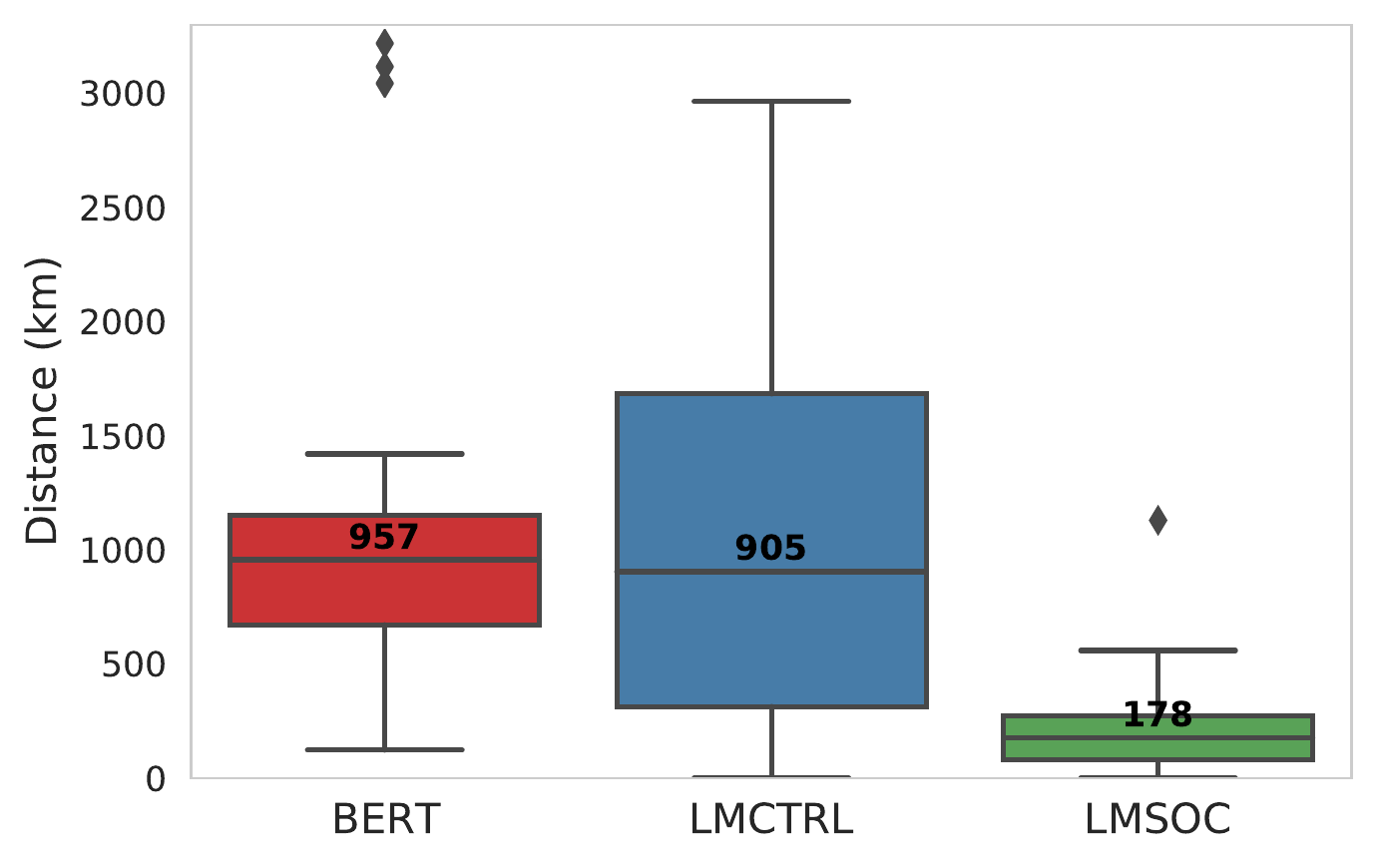}
	\caption{Descriptive statistics of the distances of the top cities from the input city predicted by various models on the \textbf{CLOSECITY} task (lower is better).  See Section \ref{subsubsec:real-eval-closecity} for details.}
\label{fig:task3}
\end{figure}

\paragraph{Results.} Table \ref{tab:task2} shows the results of our evaluation for the \textbf{STATES} and \textbf{NFL} tasks. While models that leverage social context generally perform better than \bertmethod\ on both tasks (as measured by MRR), we also observe that our model \ourmethod\ significantly outperforms \ctrlmethod\ because \ourmethod\ generalizes better to social contexts not seen during training (see Table \ref{tab:task2_qualitative} for sample predictions). 

Similar conclusions can be drawn from the results on the \textbf{CLOSECITY} task as well. Figure \ref{fig:task3} shows the summary statistics of the distances of the top city predicted by various models on the \textbf{CLOSECITY} task. Note once again, that the median distance  (from the input location) of the cities predicted by the \ourmethod\ ($178$ km) model is significantly lower than \bertmethod\ ($957$ km) and \ctrlmethod\ ($905$ km). Examining the predictions made by \ourmethod\ also suggests that \ourmethod\ is able to condition its predictions so that they align with geographical proximity better than other models considered. For example, when the input context is ``Pittsburgh'' our model prefers to predict ``Columbus (Ohio State)'' which is about $261$ km away over other major cities in the state of Pennsylvania like Philadelphia ($489$ km) and Allentown ($382$ km) thus aligning with the observation that Columbus is closer to Pittsburgh than Philadelphia and Allentown. Similarly, when the input context is ``Buffalo (NY)'', the model prefers to predict ``Toronto (Canada)'' (which is closer) over other major cities in the state of New York like Rochester or New York City. In summary, these results underscore the effectiveness of \ourmethod\ in incorporating social context.

\section{Conclusion}

We proposed a method to learn socially sensitive contextualized representations from large-scale language models.  Our method embeds social context in continuous space using graph representation algorithms and proposes a simple but effective socially sensitive pre-training approach. Our approach thus enables language models to leverage correlations between social contexts and thus generalize better to social contexts not observed in training.  More broadly, our method sets the stage for future research on incorporating new types of social contexts and enabling NLP systems like personalized predictive typing systems and entity-linking systems to better accommodate language variation.

\section*{Acknowledgments}
We would like to thank Yury Malkov, Shivam Verma, and Dan Jurafsky for comments on early drafts as well as the anonymous reviewers who suggested additions to the paper. 
\bibliographystyle{acl_natbib}
\bibliography{anthology,acl2021}

\appendix
\clearpage

\section{Data Statement}
In this section, as per recommendations outlined in \cite{bender2018data}, we describe additional details on the training data set of tweets used for the tasks described in  Section \ref{subsec:real-eval}.  

\textsc{Summary} -- To construct our training data, we obtain a random sample of $10$ million English tweets grounded in $10$ major US cities. 

\textsc{Curation Rationale} -- In particular the tweets that originated from the following 10 major cities: Los Angeles, Houston, Jacksonville, Buffalo, Philadelphia, Chicago, Columbus, Atlanta, Charlotte, Detroit. The unseen social contexts we evaluate our models are: San Diego, San Jose, San Francisco, Fresno, San Antonio, Dallas, Austin, Fort Worth, Miami, Tampa, Orlando, St. Petersburg, Rochester,	New York City, Yonkers, Syracuse, Pittsburgh, Allentown, Erie, Reading, Aurora, Naperville, Joliet, Rockford, Cleveland, Cincinnati, Toledo, Akron, Augusta, Columbus (Georgia), Macon, Savannah, Raleigh, Greensboro, Durham, Winston-Salem, Grand Rapids, Warren, Sterling Heights, Ann Arbor. 

We use this resource that lists NFL teams by state here: \url{https://state.1keydata.com/nfl-teams-by-state.php} as a reference for the team names of NFL teams for various states.

The rationale for this setup was primarily driven by our aim to evaluate our proposed approach effectively in the simplest possible setting and ease of experiment design. In addition, the size of the data acquired was also influenced by constraints on compute available for training, and time available for experimentation. 

\textsc{Language variety} -- The data was collected using Twitter API around January, 2021. The tweets were restricted to English only. More fine-grained information is not available. 

\textsc{Speaker demographic} -- Demographic information of the users is not available for this data. One would expect the demographic information to be similar to the demographics of Twitter users in the USA around January $2021$.

\textsc{Annotator demographic} -- Not applicable. Our raw dataset does not require any human annotations. 

\textsc{Text Characteristics} -- In general, tweets tend to be short, informal text. The maximum length of a tweet is at-most $280$ characters. The intended audience of a tweet is mostly other Twitter users. 

\section{Modeling Social Contexts Using Node2vec}
\label{sec:graph-rep}
Here, we outline more details on our approach to modeling social contexts. We reiterate that one may use any approach to implement social context encoder as long as it subscribes to the input, output requirements outlined in Section \ref{sec:model}. In our work, we propose one such approach using graph representation learning algorithms. Our approach uses two steps:
\begin{enumerate}
\item Constructs a graph that encodes similarities between social contexts. This requires expertise and knowledge specific to the social context being modeled.
\item Use a graph representation algorithm to learn dense embeddings of the nodes in the graph thus encoding similarities in social context. 
\end{enumerate}

As an expedient choice, in our work we use \textsc{Node2Vec} \cite{grover2016node2vec} as the graph representation algorithm to embed nodes in the constructed graph because of its simplicity and ease of training. However, one could use more advanced methods like \textsc{GraphSage} \cite{hamilton2017inductive} which will also enable inductive learning of social context embeddings. We now discuss applications of this approach to embed time, and geographic locations.

\paragraph{Embedding Time.} To embed time as represented by chronological years, we first need to encode our intuitive understanding of similarities in time points (years). In particular, we need to encode the intuitive notion that $1902$ is more similar to $1901$ and $1903$ than $1995$. Noting that time advances forward in a linear fashion, a natural way to model similarity among years is via a simple path graph. We thus  construct a simple path graph (a linear chain) where year $y$ is connected to $y-1$ and $y+1$ (the previous year, and the next year when available). We then use \textsc{node2vec} on this simple path graph which will then yield a dense representation of each year. 

\paragraph{Embedding Geographic Location.} We assume each geographic location can be represented by its geographic co-ordinates (latitude, longitude). Intuitively, we would like embeddings of locations that are close to each other geographically to also be close in embedding space. To encode this intuition, and construct a graph that encodes this notion, we first find a suitable distance measure $d$ that computes the distance between any two geographic locations given their co-ordinates. The natural distance measure here is the geodesic distance. Given this distance measure, we can now construct a directed graph where each location is connected to its $k$-closest neighbors which can then be converted to an undirected graph over which \textsc{node2vec} can be run.

Finally, the above approach can also be generalized to embed more complicated types of social-contexts (beyond time, and locations) as long as one is able to design/engineer a distance measure $D(c_{1}, c_{2})$ between any pair of contexts $\langle c_{1}, c_{2}\rangle$.

\section{Experimental Settings and Hyperparameters}
\paragraph{Node2Vec Settings.} We embed nodes into $d=768$ dimensions the same size as that of \textsc{Bert} word piece embeddings. The walk length and number of walks is set to $5$ and $1000$ respectively. 

\paragraph{Experimental settings for Evaluation Tasks.} For pre-training language models, we use the standard parameters for masked language modeling pre-training defined by \textsc{HuggingFace} transformers \cite{wolf-etal-2020-transformers}. For the evaluation task on synthetic corpus we pre-train all models for $2000$ steps (noting that loss converges at this point). For the evaluation task on real world language data, we pretrain all of our models for $3$ epochs using a batch size of $64$. During training, we set the number of warm-up steps to $500$. For both tasks, we use the AdamW optimizer with the default initial learning rate of $0.001$ and use a weight decay of $0.01$. The training time on the synthetic corpus and the real world corpus is around $5$ minutes and $16$ hours respectively on $1$ V100 GPU with $16$GB memory. Finally a note on evaluation -- in the instance when reference answer is split into multiple tokens, we accept the highest ranked answer which matches any of these tokens. 

\section{Code and Data Availability}
Code is available at \url{https://github.com/twitter-research/lmsoc}.
\end{document}